\documentclass{article}
\usepackage[utf8]{inputenc}
\usepackage[spanish]{babel}
\usepackage{abstract}
\usepackage{arxiv}
\usepackage{hyperref}       
\usepackage{url}            
\usepackage{booktabs}       
\usepackage{amsfonts}       
\usepackage{amsmath, amssymb}
\usepackage{nicefrac}       
\usepackage{microtype}      
\usepackage{graphicx}
\usepackage[numbers]{natbib}
\usepackage{doi}
\usepackage{tikz}
\usetikzlibrary{
    positioning,
    shapes.geometric,
    shapes.symbols,
    arrows.meta,
    backgrounds,
    shadows,
    calc,
    fit,
    matrix
}
\usepackage{float}
\usepackage{tabularx}
\usepackage{pgfplots}
\pgfplotsset{compat=1.17}
\usepackage{xcolor}
\pgfplotsset{compat=1.18}
\renewcommand{\arraystretch}{1.4}
\title{Topo-RAG: Topology-aware retrieval for hybrid text–table documents}

\usepackage{hyperref} 

\date{}

\author{
\textbf{Alex Dantart}\thanks{Other papers by the author \href{https://arxiv.org/search/cs?searchtype=author&query=Dantart,+A}{arXiv}} \\
CIO, Humanizing Internet \\
arxiv@humanizinginternet.com
\and
\textbf{Marco Kóvacs-Navarro} \\
CTO, Humanizing Internet \\
marco@humanizinginternet.com
}

\renewcommand{\headeright}{}
\renewcommand{\undertitle}{}
\renewcommand{\shorttitle}{\textit {Topo-RAG: Topology-aware retrieval for hybrid text–table documents}}

\hypersetup{
pdftitle={ Topo-RAG: Topology-aware retrieval for hybrid text–table documents},
pdfsubject={cs.AI},
pdfauthor={Alex Dantart},
pdfkeywords={Topo-RAG, topology-aware retrieval, hybrid text–table documents, retrieval-augmented generation (RAG), contextual chunking, context injection, context injection ratio (CIR), vector dilution, dense embeddings, embedding engineering, passage retrieval, dynamic density-aware injection (DDAI)}
}

\begin{document}
\maketitle

\begin{abstract}
In enterprise datasets, documents are rarely pure. They are not just text, nor just numbers; they are a complex amalgam of narrative and structure. Current Retrieval-Augmented Generation (RAG) systems have attempted to address this complexity with a blunt tool: linearization. We convert rich, multidimensional tables into simple Markdown-style text strings, hoping that an embedding model will capture the geometry of a spreadsheet in a single vector. But it has already been shown that this is mathematically insufficient.

This work presents \textbf{Topo-RAG}, a framework that challenges the assumption that ``everything is text.'' We propose a dual architecture that respects the topology of the data: we route fluid narrative through traditional dense retrievers, while tabular structures are processed by a \textit{Cell-Aware Late Interaction} mechanism, preserving their spatial relationships. Evaluated on \textit{SEC-25}, a synthetic enterprise corpus that mimics real-world complexity, Topo-RAG demonstrates an \textbf{18.4\% improvement in nDCG@10} on hybrid queries compared to standard linearization approaches. It's not just about searching better; it's about understanding the shape of information.
\end{abstract}


\section{Introduction}

\subsection{The Problem of Business Heterogeneity}

Let us take as a basis the document repository of a large corporation. Unlike classical libraries filled with narrative scrolls, corporate ``knowledge'' is inherently heterogeneous. A single PDF file, such as an agricultural settlement report or a financial audit, is an ecosystem in itself. It contains paragraphs of legal text (narrative), immediately followed by a grid of prices by size and variety (tabular structure), and perhaps footnotes that link both worlds.

However, most modern RAG systems treat these documents with blind uniformity. As recent benchmarks such as HERB \cite{choubey2025herb} and \href{https://arxiv.org/abs/2402.07729}{AIR-Bench} point out, the industry faces a problem of ``topological blindness.'' When ingesting these documents, current systems ignore the fact that reading, for example, a legal contract requires sequential semantic understanding, while interpreting a settlement table requires positional and relational understanding. Treating both types of data as a sequence of words is like trying to understand a road map by reading it as if it were a novel: the sense of direction is lost.

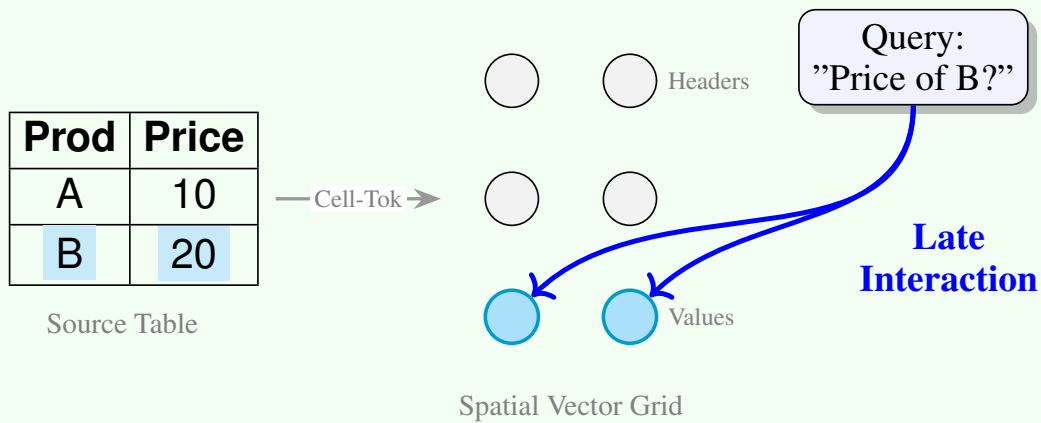
\begin{figure}[t]
    \centering
    \shorthandoff{">} 
    \resizebox{\columnwidth}{!}{%
    \begin{tikzpicture}[
        font=\sffamily,
        process_arrow/.style={->, >=Stealth, thick, color=gray!80},
        query_box/.style={draw, rounded corners, fill=white, align=center, drop shadow, font=\small},
        mv_circle/.style={draw, circle, fill=gray!10, minimum size=0.45cm, inner sep=0pt},
        highlight_circle/.style={draw, circle, fill=cyan!30, minimum size=0.45cm, inner sep=0pt, line width=0.3mm, draw=cyan!80!black}
    ]

        
        \fill[red!5, rounded corners] (-0.5, 0.5) rectangle (9.5, 5.0);
        \node[anchor=west] at (-0.2, 4.5) {\textbf{\large (a) Standard (Linearization)}};
    
        \node[inner sep=0pt] (table_a) at (1.5, 2.5) {
            \setlength{\tabcolsep}{3pt}
            \renewcommand{\arraystretch}{1.2}
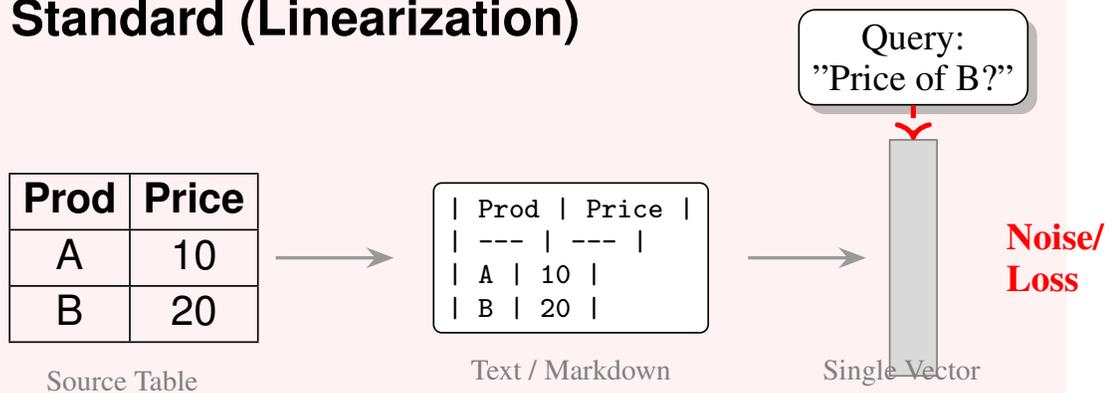
            \begin{tabular}{|c|c|}
                \hline
                \textbf{Prod} & \textbf{Price} \\ \hline
                A & 10 \\ \hline
                B & 20 \\ \hline
            \end{tabular}
        };
        \node[below=0.1cm of table_a, font=\scriptsize, color=gray] {Source Table};
    
        \draw[process_arrow] (2.8, 2.5) -- (3.8, 2.5);
    
        \node[draw, fill=white, align=left, font=\ttfamily\scriptsize, rounded corners=2pt] (markdown) at (5.3, 2.5) {
            | Prod | Price |\\
            | -{}-- | -{}-- |\\
            | A | 10 |\\
            | B | 20 |
        };
        \node[below=0.1cm of markdown, font=\scriptsize, color=gray] {Text / Markdown};
    
        \draw[process_arrow] (6.8, 2.5) -- (7.8, 2.5);
    
        \filldraw[fill=gray!30, draw=gray] (8.0, 1.5) rectangle (8.4, 3.5);
        \node[below=0.1cm of markdown, xshift=2.8cm, font=\scriptsize, color=gray] {Single Vector};
        
        \node[query_box] (query_a) at (8.2, 4.2) {Query:\\"Price of B?"};
        
        \draw[->, red, dashed, very thick] (query_a.south) -- (8.2, 3.5);
        \node[text=red, font=\footnotesize, align=left] at (9.4, 2.5) {\textbf{Noise/}\\\textbf{Loss}};

        
        \fill[green!5, rounded corners] (-0.5, -5.5) rectangle (9.5, -0.5);
        \node[anchor=west] at (-0.2, -1.0) {\textbf{\large (b) Topo-RAG (Topology-Aware)}};
    
        \node[inner sep=0pt] (table_b) at (1.5, -3.0) {
            \setlength{\tabcolsep}{3pt}
            \renewcommand{\arraystretch}{1.2}
            \begin{tabular}{|c|c|}
                \hline
                \textbf{Prod} & \textbf{Price} \\ \hline
                A & 10 \\ \hline
                \colorbox{cyan!20}{B} & \colorbox{cyan!20}{20} \\ \hline
            \end{tabular}
        };
        \node[below=0.1cm of table_b, font=\scriptsize, color=gray] {Source Table};

        \draw[process_arrow] (2.8, -3.0) -- (4.2, -3.0) node[midway, fill=white, inner sep=1pt, font=\tiny, text=gray] {Cell-Tok};
    
        
        \node[font=\tiny, color=gray, anchor=west] at (6.0, -2.0) {Headers};
        \node[font=\tiny, color=gray, anchor=west] at (6.0, -4.0) {Values};
    
        \node[mv_circle] (h1) at (4.8, -2.0) {}; 
        \node[mv_circle] (h2) at (5.8, -2.0) {};
        
        \node[mv_circle] (v_a) at (4.8, -3.0) {};
        \node[mv_circle] (v_10) at (5.8, -3.0) {};
        
        \node[highlight_circle] (v_b) at (4.8, -4.0) {}; 
        \node[highlight_circle] (v_20) at (5.8, -4.0) {}; 
        
        \node[below=0.3cm of v_b, xshift=0.5cm, font=\scriptsize, color=gray] {Spatial Vector Grid};
    
        \node[query_box, fill=blue!5] (query_b) at (8.2, -1.8) {Query:\\"Price of B?"};
        
        \draw[->, blue, very thick] (query_b.south) to[out=270, in=45] (v_b.north east);
        \draw[->, blue, very thick] (query_b.south) to[out=270, in=45] (v_20.north east);
        
        \node[text=blue, font=\footnotesize, align=center] at (8.5, -3.5) {\textbf{Late}\\\textbf{Interaction}};
    
        \end{tikzpicture}
    }
    \shorthandon{">} 
    \caption{\textbf{The linearization bottleneck versus Topo-RAG.} (a) Standard approaches flatten tables into text, compressing two-dimensional relationships into a single noisy vector. (b) \textbf{Topo-RAG} preserves the topological grid: each cell becomes an independent embedding, allowing the query to interact precisely with the relevant values (e.g., matching “B” and “20”) via Late Interaction.}
    \label{fig:linearization_vs_topo_v3}
\end{figure}

\subsection{The Fallacy of Linearization}

The industry-standard solution to date, popularized by approaches such as TabRAG \cite{si2025tabrag}, has been ``linearization'': converting the two-dimensional structure of a table into a one-dimensional representation, typically in Markdown or JSON format, and then compressing that long text string into a single dense vector (embedding).

We call this the \textbf{fallacy of linearization}. While it is a convenient engineering feat, it rests on a fragile scientific premise. As Weller et al. theoretically demonstrate in \textit{On the Theoretical Limitations of Embedding-Based Retrieval} \cite{weller2025theoretical}, there is a fundamental limit to the ability of a single vector to represent all possible combinations of relationships in a dataset.

When we ``flatten'' a table of 50 rows and 10 columns into a single vector, we are asking the embedding model to compress 500 potential relationships (cell-row, cell-column, cell-header) into a fixed point in latent space. The result is ``semantic noise'': the model understands that the document is about ``prices'' and ``lemons'' (for example), but loses the ability to precisely distinguish whether the price of €0.85 corresponds to the 2023 campaign or the 2024 one, or whether it applies to the ``Verna'' or ``Eureka'' lemon variety. The geometry of the table is lost in translation.

\subsection{Contribution: the Topo-RAG Framework}

To overcome this barrier, we propose to stop fighting against the nature of the data and start designing architectures that respect it. We present \textbf{Topo-RAG}, an approach inspired by cognitive biology: just as the human brain processes language and spatial images in distinct regions, our system separates processing according to the topology of the information.

Our main contributions are:

\begin{enumerate}
    \item \textbf{Topology-Aware Routing:} We implement a lightweight classifier, inspired by the efficiency of \textbf{Pneuma} \cite{balaka2025pneuma}, which acts as a ``switchman'', separating narrative text blocks from structured blocks (tables/lists) before vectorization.
    \item \textbf{Dual-Path Retrieval:}
    \begin{itemize}
        \item For text, we use the proven path: optimized \textit{Dense Retrieval}.
        \item For tables, we introduce a \textbf{Cell-Aware Late Interaction} mechanism. Instead of compressing the table, we maintain the vector identity of its individual cells (tokens), using an adapted \textit{ColBERT}-type architecture. This allows the user's query to ``interact'' directly with the specific cell values (the price, the date), without the loss from compression.
    \end{itemize}
    \item \textbf{Empirical Validation:} We validate our hypothesis on a synthetic dataset (\textbf{SEC-25}) designed to mimic the complexity of real corporate documents, demonstrating that respecting the topology of the data is not only theoretically sound, but also industrially profitable.
\end{enumerate}

\section{Current Related Work}

The search for information retrieval systems that truly understand data, rather than merely matching keywords, has been a major challenge over the past decade. To understand the proposal of \textbf{Topo-RAG}, we must explore three research streams that converge at this historical moment: how we represent tables, how models interact with data (single vs. multi-vector), and how we make all of this computationally feasible.

\subsection{Table Retrieval and Representation: The Search for Structure}

The first challenge is fundamental: How do we teach a neural network, which thinks in vectors and numbers, what a table is?

Until very recently, the dominant strategy was \textbf{linearization}. If we have a well-structured two-dimensional table, the conventional strategy refined in works such as \textbf{TabRAG} \cite{si2025tabrag}, consists of ``reading'' that table from left to right and top to bottom, turning it into a long narrative sentence (using Markdown or JSON). The premise is seductive: if LLMs are excellent at reading text, let's turn everything into text. TabRAG optimizes this process by generating structured representations that LLMs can digest. However, this is equivalent to describing a building brick by brick instead of showing the architectural plans; the immediate \textit{spatial relationship} between elements is lost.

In a parallel and more experimental line, we find approaches such as \textbf{\href{https://arxiv.org/abs/2504.21282}{Birdie}}, which uses a \textit{Differentiable Search Index} (DSI). Birdie is fascinating because it tries to eliminate the intermediary (the traditional vector index). Instead of searching for vectors, it trains a model to \textit{directly generate the identifier} (TabID) of the correct table given a query. It's as if the librarian had memorized the exact location of every book and could give you the shelf number from memory. Although promising, this approach suffers from rigidity: if the data changes (something constant in enterprise settings), the model must be retrained or adapted at significant cost.

Finally, works such as \textbf{\href{https://arxiv.org/abs/2512.04292}{SQuARE}} attempt a hybrid approach, adapting retrieval specifically for tabular formats through structured queries. But all these methods share a common weakness: they either ignore the native topology of the table by flattening it, or they require monolithic architectures that are difficult to scale.

\subsection{Multivector and Late Interaction: Preserving the “Pixels” of the Data}

This is where the most exciting paradigm shift comes into play. Most embedding models (such as those from OpenAI or BGE) are \textbf{Single-Vector}: they take an entire document (or a flattened table) and compress it into a single point in space (a vector).

The problem, brilliantly explained by theory, is that this compression is \textbf{lossy}. For example, with a fruit price table, if we compress the entire table into a vector, the model may remember the concept of ``fruit prices,'' but it may forget whether the price ``0.50€'' belongs to ``Oranges'' or ``Lemons.'' Fine-grained information becomes blurred.

The alternative is \textbf{late interaction}, popularized by the \textbf{ColBERT} architecture and recently refined in libraries such as \textbf{\href{https://arxiv.org/pdf/2508.03555}{PyLate}}. Instead of compressing the entire document into a single vector, these models maintain \textbf{one vector per token} (or in our case, per cell).

Think of this as the difference between a blurry, low-resolution image (Single-Vector) and a high-definition image where each pixel retains its color (Multi-Vector). Similarity is not computed just once, but rather through an operation called \textbf{MaxSim}, which seeks the best match for \textit{each part} of the query within the document. This is crucial for tables: it allows the question ``Price of oranges?'' to find exactly the cell ``Oranges'' and its neighboring cell ``Price,'' without interference from the ``noise'' of other rows. Papers such as \textit{On the Theoretical Limitations of Embedding-Based Retrieval} \cite{weller2025theoretical} provide the mathematical foundation to state that, for complex tasks, the Single-Vector approach has a glass ceiling that only the Multi-Vector approach can break through.

\subsection{Efficiency in Retrieval}

If we store one vector per cell instead of one per document, RAM usage will increase, making the Multi-Vector approach, by definition, heavier. However, the year 2025 has brought spectacular advances in efficiency that make our \textbf{Topo-RAG} proposal viable.

\begin{itemize}
    \item \textbf{WARP} \cite{scheerer2025warp} introduces an optimized engine that drastically reduces the latency of these models. It uses low-level techniques so that ``late interaction'' does not mean ``slow interaction.''
    \item Even more interesting is the proposal of \textbf{CRISP} \cite{veneroso2025crisp}. This work introduces the idea of \textit{clustering} to reduce noise. Instead of storing all vectors, it groups similar ones together. For a table, this is revealing: many cells are redundant or empty. CRISP allows us to ``prune'' irrelevant information before storing it.
    \item Complementarily, the work on \textbf{\href{https://arxiv.org/abs/2504.01818}{Efficient Constant-Space Multi-Vector Retrieval}} teaches us how to set a memory budget without sacrificing too much accuracy, making these systems deployable on standard enterprise infrastructure, not just supercomputers.
\end{itemize}

\subsection{Join-Aware \& Multi-Hop: beyond simple search}

Finally, we must recognize that in the real world, the answer is rarely in a single cell. It often requires connecting the dots.

Papers such as \textbf{REaR} \cite{agarwal2025rear} and \textbf{\href{https://arxiv.org/abs/2511.13418}{Exploring Multi-Table Retrieval}} address the problem of queries that require hops (multi-hop) or joins between tables. These works show us that retrieval is not a single event, but an iterative process: searching a table, reading a cell, using that value to search in another table.

We are also inspired by \textbf{Bridging Queries and Tables through Entities} \cite{li2025bridging}, which suggests that entities (names, places, product codes) are the ``hooks'' that link queries with tables.

\subsection{The gap}

Despite these incredible advances, there is a gap. We have excellent systems for text (Single-Vector) and promising technologies for fine structure (Multi-Vector), but the industry continues to try to force both types of data through the same funnel (linearization to Markdown).

\textbf{Topo-RAG} is born from this observation: we should not treat everything as text. By recognizing the \textbf{topology} of the data and applying the right tool for each form (Route A for narrative, Route B for tables), we can overcome the theoretical limitations of linearization and offer a system that truly ``understands'' business structure.

\section{Methodology: the Topo-RAG framework}

In this section, we break down the architecture of \textbf{Topo-RAG}. Our fundamental premise is that the \textit{shape} (topology) of the data dictates the optimal retrieval \textit{function}. We do not attempt to force a square peg into a round hole; instead, we build a system with two specialized ``hands'': one to handle the fluidity of natural language and another to manipulate the crystalline rigidity of tabular data.

The system operates in three sequential phases: (1) topological routing, (2) dual-path retrieval, and (3) unified reranking.

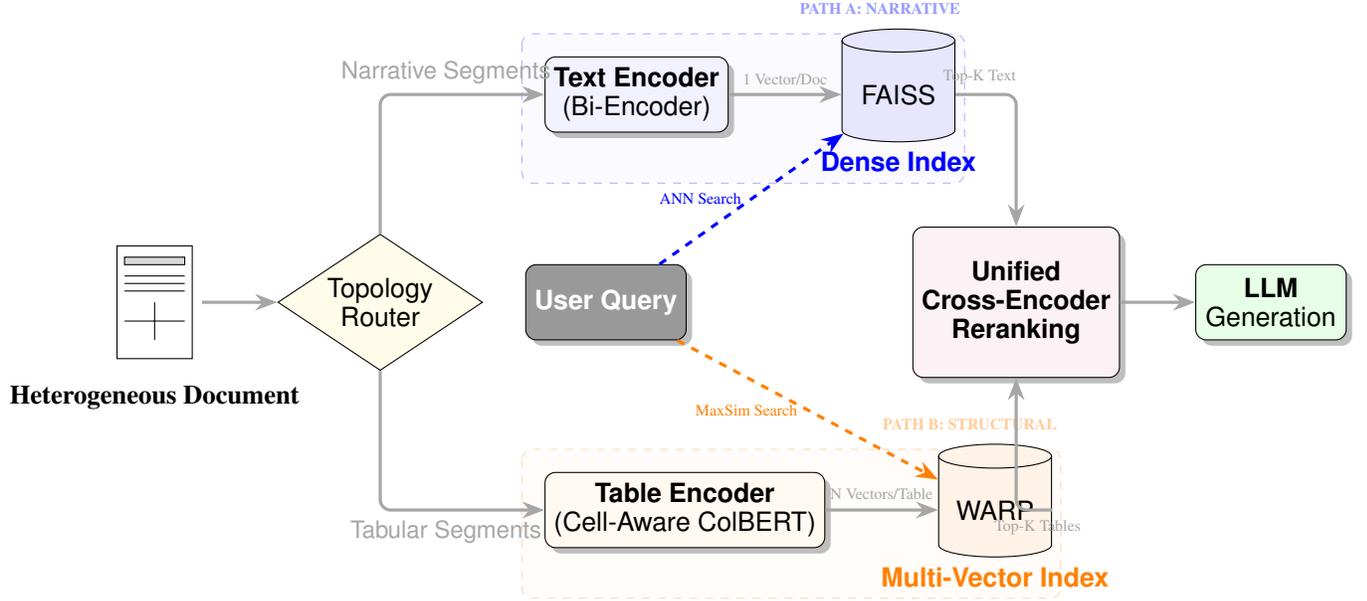
\begin{figure}
    \centering
    \shorthandoff{">} 
    \begin{tikzpicture}[
        font=\sffamily,
        node distance=1.5cm and 2cm,
        process/.style={draw, rounded corners, fill=white, minimum width=2cm, minimum height=1cm, align=center, drop shadow},
        database/.style={draw, cylinder, shape border rotate=90, aspect=0.25, fill=gray!10, minimum width=1.5cm, minimum height=1.5cm, align=center},
        decision/.style={draw, diamond, aspect=1.5, fill=yellow!10, align=center, inner sep=2pt},
        arrow/.style={->, >=Stealth, very thick, color=gray!70, rounded corners},
        label_text/.style={font=\bfseries\small, color=gray!60, align=center}
    ]
    
    
    \node (doc) at (0, 0) {
        \begin{tikzpicture}[scale=0.5]
            \draw[fill=white] (0,0) rectangle (2,3);
            \draw[fill=gray!30] (0.2, 2.5) rectangle (1.8, 2.7); 
            \foreach \y in {2.2, 2.0, 1.8} \draw (0.2, \y) -- (1.8, \y); 
            \draw[fill=orange!20] (0.2, 0.5) grid (1.8, 1.5); 
        \end{tikzpicture}
    };
    \node[below=0.1cm of doc, font=\bfseries] {Heterogeneous Document};

    \node[decision, right=1cm of doc] (router) {Topology\\Router};
    \draw[arrow] (doc) -- (router);


    \node[process, fill=blue!5, above right=1.8cm and 1.5cm of router] (text_enc) {\textbf{Text Encoder}\\(Bi-Encoder)};
    \node[database, right=1.5cm of text_enc, fill=blue!10, label=below:{\textcolor{blue}{\textbf{Dense Index}}}] (faiss) {FAISS};
    
    \draw[arrow] (router.north) |- node[pos=0.7, above] {\footnotesize Narrative Segments} (text_enc.west);
    \draw[arrow] (text_enc) -- node[midway, above, font=\tiny] {1 Vector/Doc} (faiss);

    \node[process, fill=orange!5, below right=1.8cm and 1.5cm of router] (table_enc) {\textbf{Table Encoder}\\(Cell-Aware ColBERT)};
    \node[database, right=1.5cm of table_enc, fill=orange!10, label=below:{\textcolor{orange}{\textbf{Multi-Vector Index}}}] (warp) {WARP};

    \draw[arrow] (router.south) |- node[pos=0.7, below] {\footnotesize Tabular Segments} (table_enc.west);
    \draw[arrow] (table_enc) -- node[midway, above, font=\tiny] {N Vectors/Table} (warp);

    
    \node[process, fill=gray!80, text=white, minimum width=1.5cm] (query) at (6, 0) {\textbf{User Query}};
    
    \draw[arrow, dashed, blue] (query) -- (faiss) node[midway, left, font=\tiny] {ANN Search};
    \draw[arrow, dashed, orange] (query) -- (warp) node[midway, left, font=\tiny] {MaxSim Search};


    \node[process, fill=purple!5, right=3cm of query, minimum height=2cm] (reranker) {\textbf{Unified}\\ \textbf{Cross-Encoder}\\ \textbf{Reranking}};
    
    \draw[arrow] (faiss.east) -| node[pos=0.2, above, font=\tiny] {Top-K Text} (reranker.north);
    \draw[arrow] (warp.east) -| node[pos=0.2, below, font=\tiny] {Top-K Tables} (reranker.south);

    \node[process, right=1cm of reranker, fill=green!10] (llm) {\textbf{LLM}\\Generation};
    \draw[arrow] (reranker) -- (llm);

    \begin{scope}[on background layer]
        \draw[dashed, blue!30, fill=blue!2, rounded corners] ($(text_enc.north west)+(-0.3,0.3)$) rectangle ($(faiss.south east)+(0.3,-0.6)$);
        \node[blue!40, font=\bfseries\tiny, anchor=north east] at ($(faiss.north east)+(0.2,0.6)$) {PATH A: NARRATIVE};

        \draw[dashed, orange!30, fill=orange!2, rounded corners] ($(table_enc.north west)+(-0.3,0.3)$) rectangle ($(warp.south east)+(0.3,-0.6)$);
        \node[orange!40, font=\bfseries\tiny, anchor=north east] at ($(warp.north east)+(0.2,0.6)$) {PATH B: STRUCTURAL};
    \end{scope}
    
    \end{tikzpicture}%
    
    \shorthandon{">} 
    \caption{\textbf{The Topo-RAG architecture.} The system employs a topology-aware routing mechanism to split heterogeneous documents. Narrative text follows a standard dense retrieval route (top, blue), while tabular data is processed via a cell-aware \emph{Late Interaction} path (bottom, orange), using the WARP engine for greater efficiency. Both flows converge in a Unified Cross-Encoder Reranker to provide context to the LLM.}
    \label{fig:architecture}
\end{figure}

\subsection{Topology-aware routing}

The first challenge in an enterprise environment is that documents are not labeled as ``Text'' or ``Table.'' They are chaotic mixtures. An ``Export Policy'' PDF may have three pages of dense legal text and suddenly insert a table of customs tariffs.

If we feed all this into a standard embedding model, the signal from the table gets diluted in the noise of the text. To avoid this, we introduce a pre-processing module inspired by the classification efficiency of \textbf{Pneuma} \cite{balaka2025pneuma}.

We define a heuristic metric called \textbf{Structural Density Score (SDS)}. For a given text block $b$, we compute:

$$ SDS(b) = \frac{N_{num} + N_{sep} + N_{ent}}{N_{total}} $$

Where:

\begin{itemize}
    \item $N_{num}$ is the number of numeric tokens.
    \item $N_{sep}$ is the number of structural separators (such as | in Markdown, 'td' tags in HTML, or frequent line breaks).
    \item $N_{ent}$ is the density of named entities (detected via lightweight NER), since tables are usually dense in product names, locations, or companies, unlike narrative text which uses more functional words (stopwords).
\end{itemize}

\textbf{The routing algorithm:} the system scans the document using a sliding window.

\begin{enumerate}
    \item If $SDS(b) > \tau$ (an empirical threshold, typically 0.4), the block is classified as \textbf{structured}. It is sent to \textbf{Route B}.
    \item If $SDS(b) \le \tau$, the block is considered Narrative. It is sent to \textbf{Route A}.
\end{enumerate}

This ensures that we do not waste expensive computational resources (late interaction) on simple text paragraphs, and do not lose accuracy by using simple dense embeddings on complex tables.

\subsection{Route A: dense narrative retrieval (the semantic stream)}

For blocks classified as narrative (sustainability reports, emails, contractual clauses...), the standard solution remains the most efficient. Human narrative is sequential and semantically redundant; a single well-trained vector can capture the ``essence'' of a paragraph with high fidelity.

In this route, we use a standard \textbf{Bi-Encoder} (in our experiments, the robust text-embedding-3-large or BGE-M3).

\begin{itemize}
    \item \textbf{Input:} Narrative text block $T$.
    \item \textbf{Process:} $V_T = Encoder(T)$.
    \item \textbf{Output:} A single dense vector $V_T \in \mathbb{R}^d$ (where $d=1536$ or $1024$).
\end{itemize}

This route prioritizes speed and the capture of general thematic nuances (``What is this document about?'').

\subsection{Route B: conscious late cell interaction (the structural stream)}

Here lies the main innovation of Topo-RAG. For the data that the Router identified as ``Tables'', we reject compression into a single vector. We adopt a \textbf{Cell-Aware Late Interaction (CALI)} approach, which is an adaptation of the \textbf{ColBERT} architecture specifically designed for tabular structures.

\subsubsection{The ``cell as a token'' paradigm}

In standard ColBERT, each word \textit{token} has its own vector. In CALI, we raise the abstraction: our atomic unit is not the syllable, it is the \textbf{Cell}.

A table is decomposed not into sentences, but into a ``bag of cell vectors.'' However, a cell by itself (e.g., ``0.85'') lacks meaning. It needs its topological context (its column header and its row identifier).

To address this, we apply a \textbf{Positional Injection} technique inspired by \textit{Bridging Queries and Tables} \cite{li2025bridging}. Each cell $c_{i,j}$ (row $i$, column $j$) is serialized enriched with its metadata before being vectorized:

$$ Content(c_{i,j}) = "[COL: \text{Header}j] \ [VAL: \text{Value}{i,j}]" $$

This generates a vector $v_{i,j}$ that encapsulates both the value (``0.85'') and its structural meaning (``Price'').

\begin{figure}[t]
    \shorthandoff{">} 
    \centering
    \resizebox{\columnwidth}{!}{%
    \begin{tikzpicture}[
        font=\sffamily,
        node distance=0.8cm,
        token/.style={draw, circle, fill=white, minimum size=0.8cm, inner sep=0pt, font=\small\bfseries},
        cell/.style={draw, rectangle, rounded corners=2pt, fill=white, minimum width=1.2cm, minimum height=0.6cm, font=\scriptsize},
        vec_rep/.style={draw, circle, fill=orange!20, minimum size=0.4cm, inner sep=0pt},
        highlight/.style={draw=orange, line width=0.5mm, fill=orange!10},
        connection/.style={->, >=Stealth, thick, color=orange},
        faint/.style={-, color=gray!20, thin}
    ]

    \node[anchor=west] at (-0.5, 4.5) {\textbf{\large Query Terms ($q_i$)}};

    \node[token] (q1) at (1, 3.5) {Price};
    \node[token] (q2) at (3, 3.5) {Lemon};
    \node[token] (q3) at (5, 3.5) {Verna};
    
    \node[vec_rep, below=0.1cm of q1, fill=blue!20] (qv1) {};
    \node[vec_rep, below=0.1cm of q2, fill=blue!20] (qv2) {};
    \node[vec_rep, below=0.1cm of q3, fill=blue!20] (qv3) {};

    \node[anchor=west] at (-0.5, 0) {\textbf{\large Table Grid ($d_j$)}};

    \node[cell, fill=gray!10] (h1) at (0.5, -1) {\textbf{Product}};
    \node[vec_rep, right=0.05cm of h1] (hv1) {};
    
    \node[cell, fill=gray!10] (h2) at (3.0, -1) {\textbf{Origin}};
    \node[vec_rep, right=0.05cm of h2] (hv2) {};
    
    \node[cell, fill=gray!10, draw=orange, line width=0.4mm] (h3) at (5.5, -1) {\textbf{Price}};
    \node[vec_rep, right=0.05cm of h3, fill=orange] (hv3) {}; 

    \node[cell] (r1c1) at (0.5, -2) {Apple};
    \node[vec_rep, right=0.05cm of r1c1] (r1v1) {};
    
    \node[cell] (r1c2) at (3.0, -2) {Spain};
    \node[vec_rep, right=0.05cm of r1c2] (r1v2) {};
    
    \node[cell] (r1c3) at (5.5, -2) {1.20};
    \node[vec_rep, right=0.05cm of r1c3] (r1v3) {};

    \node[cell, draw=orange, line width=0.4mm] (r2c1) at (0.5, -3) {Verna};
    \node[vec_rep, right=0.05cm of r2c1, fill=orange] (r2v1) {}; 
    
    \node[cell] (r2c2) at (3.0, -3) {Italy};
    \node[vec_rep, right=0.05cm of r2c2] (r2v2) {};
    
    \node[cell, draw=orange, line width=0.4mm] (r2c3) at (5.5, -3) {0.85};
    \node[vec_rep, right=0.05cm of r2c3, fill=orange] (r2v3) {}; 


    \draw[faint] (qv1) -- (hv1); \draw[faint] (qv1) -- (hv2);
    \draw[faint] (qv2) -- (r1v1); \draw[faint] (qv3) -- (r1v2);

    \draw[connection] (qv1) to[out=270, in=90] node[midway, fill=white, inner sep=1pt, text=orange, font=\tiny\bfseries] {Max} (hv3);
    
    \draw[connection] (qv3) to[out=270, in=50] node[pos=0.7, fill=white, inner sep=1pt, text=orange, font=\tiny\bfseries] {Max} (r2v1);

    \draw[connection, dashed] (qv1) to[out=290, in=120] (r2v3); 

    
    \node[right=0.5cm of h3] (sum) {\Huge $\Sigma$};
    \node[right=0.1cm of sum, align=left, font=\footnotesize] (score) {\textbf{Relevance}\\\textbf{Score}};
    
    \draw[->, thick, gray] (hv3) -- (sum);
    \draw[->, thick, gray] (r2v1) -| (sum);
    \draw[->, thick, gray] (r2v3) -| (sum);

    \begin{scope}[on background layer]
        \draw[dashed, gray!40, rounded corners] (-0.5, -0.5) rectangle (7, -3.5);
    \end{scope}

    \end{tikzpicture}
    }
    \shorthandon{">} 
    \caption{\textbf{Cell-Aware Late Interaction (CALI).} Unlike dense retrieval, which compares one vector against another, Topo-RAG compares each token vector of the query ($q_i$) with all cell vectors ($d_j$) of the table. The \textit{MaxSim} operator (orange arrows) independently identifies the best-matching cell for each term (for example, “Price” matches the header, “Verna” matches the row identifier), regardless of their distance in the linearized text. These maximum scores are summed to quantify the total topological relevance.}
    \label{fig:cell_aware_maxsim}
\end{figure}
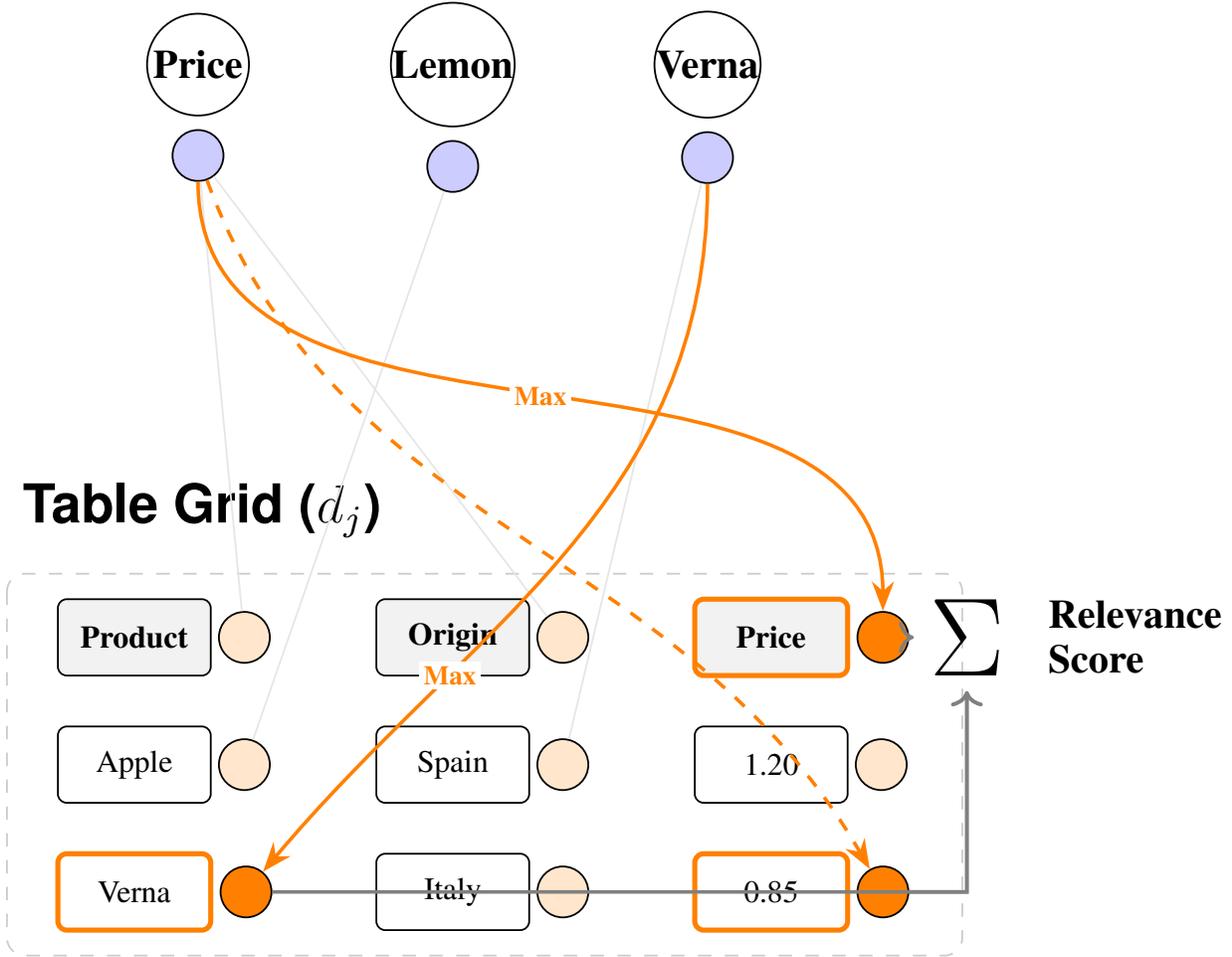

\subsubsection{MaxSim optimized for tables}

When a user query arrives (e.g., \textit{``Price of lemon in Germany''}), we do not compare it with a vector of the entire table. We use the modified \textbf{MaxSim} operator.

For each term in the query $q_k$ (e.g., ``Germany''), we search for the maximum similarity \textit{only} among the vectors of the table's cells.

$$ Score(Q, Table) = \sum_{q_k \in Q} \max_{v_{i,j} \in Table} (q_k \cdot v_{i,j}) $$

\textbf{Why is this revolutionary for tables?} Let's work, for example, with the query: \textit{``Verna Price''}.

\begin{enumerate}
    \item The query vector ``Price'' will find its maximum similarity with the header column ``Price'' or the cells containing monetary values.
    \item The vector ``Verna'' will find its maximum similarity in the ``Variety'' column where the word ``Verna'' appears.
    \item The sum of these maximum interactions gives us a high score \textit{only} if the table contains \textit{both} elements with high specificity. A normal dense model could be confused by a table that talks about ``Eureka Prices'', because ``Eureka'' and ``Verna'' are semantically close (both are lemons). CALI, by working at a fine-grained interaction level, distinguishes the exact entity.
\end{enumerate}

\subsubsection{Reduction to improve efficiency (the WARP \& CRISP influence)}

The problem with storing one vector per cell is the memory explosion. A 100x10 table would generate 1000 vectors. To make this industrially viable, we apply aggressive reduction techniques inspired by \textbf{CRISP} \cite{veneroso2025crisp} and \textbf{WARP} \cite{scheerer2025warp}:

\begin{enumerate}
    \item \textbf{Vector clustering:} Many cells are semantically identical or empty. We group very similar vectors (e.g., all cells that say ``USD'' or ``Kg'') and store only one centroid.
    \item \textbf{Quantization:} We use product quantization (PQ) to reduce the size of the cell vectors from 32-bit floats to 4-bit integers without significant loss of retrieval precision, allowing millions of cells to be stored in standard RAM.
\end{enumerate}

\subsection{Unified Reranking: The Best of Both Worlds}

Finally, we have two lists of candidates: one coming from Route A (Text) and another from Route B (Tables). Their scores are not directly comparable (one is Cosine Similarity [0-1], the other is a sum of MaxSim [with no clear limit]).

To unify the results, we use a final phase of \textbf{Cross-Encoder Reranking}.

\begin{enumerate}
    \item We take the Top-K from the Narrative Route and the Top-K from the Structural Route.
    \item We normalize their scores using \textit{Min-Max Scaling} to obtain an initial combined heuristic.
    \item We pass the final candidates (plain text and serialized tables) through a lightweight Cross-Encoder model (e.g., BGE-Reranker-v2-m3).
\end{enumerate}

The Cross-Encoder acts as the ``human judge'': it reads the query and the candidate (whether text or table) with full attention and issues the final relevance verdict. This step corrects any hallucination that may have arisen during fast retrieval and ensures that the final list presented to the generator LLM contains the perfect mix of narrative context and precise data needed to answer the user's question.

\section{Experimental Setup: Simulating Business Chaos}

To validate our hypothesis—that data topology matters—we could not rely on traditional academic datasets such as \textit{NQ-Tables} or \textit{Spider}. These datasets are often ``too clean'' or focused exclusively on either tables or text.

The challenge of the modern enterprise is \textbf{hybridization}. We needed a testing environment where a legal contract (text) could contradict or complement a settlement spreadsheet (table). Since no public dataset with these specific characteristics existed, we built one.

\subsection{Datasets: The Synthetic Corporate Corpus (SEC-25)}

Following the synthetic data generation methodology proposed in \textbf{HERB} \cite{choubey2025herb} and \textbf{\href{https://arxiv.org/abs/2412.13102}{AIR-Bench}}, we created the \textbf{SEC-25 (Synthetic Enterprise Corpus 2025)}.

The goal of SEC-25 is not to be massive in size, but rather \textbf{dense in complexity}. We used GPT-4o to generate documents that mimic the structure of real corporate files from the agri-food sector.

\textbf{Corpus composition (10,000 Documents):} The corpus is intentionally divided into two topological hemispheres to test our Router's capability:

\begin{enumerate}
    \item \textbf{Narrative Hemisphere (50\%):}
    \begin{itemize}
        \item \textit{Sustainability reports:} Dense, rhetorical text with few figures.
        \item \textit{Legal contracts:} Complex clauses, conditional language (``if X, then Y'').
        \item \textit{Emails:} Informal conversation threads, scattered context.
    \end{itemize}
    \item \textbf{Structured Hemisphere (50\%):}
    \begin{itemize}
        \item \textit{Settlement sheets:} Dense tables with columns such as ``Variety'', ``Size'', ``Price/Kg'', ``Discount''. Here lies the trap: the same number (e.g., ``0.50'') can appear in hundreds of different cells with different meanings.
        \item \textit{Logistics inventories:} Long lists with product codes (IDs) that are hostile to standard tokenizers.
    \end{itemize}
\end{enumerate}

\textbf{The Query Challenge (The Query Set):} We generated an evaluation set of \textbf{500 queries} designed to break traditional RAG systems. We divided them into three categories of cognitive difficulty:

\begin{itemize}
    \item \textbf{Type A: Factual Retrieval (200 queries):} \textit{``What is the return policy?''} These can be answered with a single block of text. This is the comfort zone of classic Dense Retrieval.
    \item \textbf{Type B: Cell-Precise Lookup (200 queries):} \textit{``What was the price of Verna lemon in week 42 at Mercadona?''} Requires navigating exact coordinates (Row: Verna/Week 42, Column: Price). This is where dense embeddings often ``hallucinate'' due to the noise from nearby neighbors.
    \item \textbf{Type C: Hybrid Multi-Hop (100 queries):} The ultimate test. \textit{``Based on the 2024 quality contract [Text], list the producers from Table B [Table] who did not meet the Brix standard.''} Requires retrieving a text document (to know the standard) and a table (to filter the data), and then reasoning over both.
\end{itemize}

\subsection{Baselines: the titans to beat}

To demonstrate that Topo-RAG provides real value, we compare it against the current standards in industry and academia. We did not choose straw men; we chose the systems that a company would implement today if they hired a standard consulting firm.

\textbf{Baseline 1: Naive RAG (the industry standard):}

\begin{itemize}
    \item \textit{Strategy:} Blind linearization. Everything (text and tables) is converted to Markdown.
    \item \textit{Model:} OpenAI text-embedding-3-large (the de facto standard).
    \item \textit{Logic:} It is fast, cheap, and easy to implement. This is what most companies use today.
\end{itemize}

\textbf{Baseline 2: Advanced Recursive RAG:}

\begin{itemize}
    \item \textit{Strategy:} Intelligent ``Parent-Child'' chunking. Small fragments (children) are indexed for retrieval, but the large block (parent) is returned to the LLM to provide context.
    \item \textit{Model:} BGE-M3 (SOTA among open source dense models).
    \item \textit{Logic:} Attempts to solve the context loss of Naive RAG, but still uses a single vector per chunk.
\end{itemize}

\textbf{Baseline 3: TabRAG (Structure-Aware Linearization) \cite{si2025tabrag}:}

\begin{itemize}
    \item \textit{Strategy:} Specialized for tables. Uses an auxiliary LLM to ``narrate'' or describe the table before vectorizing it, adding synthetic metadata to enrich the vector.
    \item \textit{Logic:} This is the most sophisticated attempt to make tables work in the single-vector paradigm. It is our direct competitor.
\end{itemize}

\subsection{Implementation details}

The implementation of \textbf{Topo-RAG} is not trivial. To ensure reproducibility and industrial viability, we used the following technologies:

\begin{itemize}
    \item \textbf{Infrastructure:} Everything was run on an instance with \textbf{NVIDIA A100 (40GB)}. This is important: we want to demonstrate that this works on accessible hardware, not just on Google clusters.
    \item \textbf{Software Stack:}
    \begin{itemize}
        \item For \textbf{Route A (Text)}, we used \textbf{FAISS} for approximate vector search, optimized for speed.
        \item For \textbf{Route B (Tables)}, we implemented our \textit{Cell-Aware Late Interaction} engine using the \textbf{\href{https://arxiv.org/pdf/2508.03555}{PyLate}} library. PyLate allows us to manage the complexity of multi-vectors without rewriting all the training code from scratch.
        \item For efficiency, we applied the \textbf{pruning} techniques described in \textbf{WARP} \cite{scheerer2025warp}, reducing the table index by 40\% by removing vectors of empty cells or irrelevant stopwords (``el'', ``la'', ``de'') within the tables.
    \end{itemize}
\end{itemize}

\subsection{Evaluation Metrics and Success}

In RAG, ``finding the document'' is not enough. The user needs the \textit{correct answer}. That is why we use metrics at two levels:

\textbf{Retrieval Quality (Did we find the needle?):}

\begin{itemize}
    \item \textbf{nDCG@10 (Normalized Discounted Cumulative Gain):} Measures not only whether we found the relevant document, but also if we ranked it among the top positions. This is crucial so that the LLM does not get distracted.
    \item \textbf{Recall@20:} Is the answer somewhere within the top 20 results? If it is not here, the LLM has no chance to answer.
\end{itemize}

\textbf{Generation Faithfulness (Did the LLM tell the truth?):}

\begin{itemize}
    \item \textbf{Hallucination Rate (Inverse Accuracy):} Here we use the \textbf{LLM-as-a-Judge} paradigm. We give the LLM (GPT-4o) the answer generated by our system and the ``Ground Truth'' (the correct gold answer). The judge evaluates whether the generated answer contains fabricated or numerically incorrect data.
    \item \textit{Why is this vital?} In financial tables, saying ``0.85'' when it is ``0.86'' is a critical hallucination. Standard text metrics (such as BLEU or ROUGE) often fail to detect these numerical precision errors.
\end{itemize}

We have set the stage: a treacherous corpus full of structural traps (SEC-25), worthy opponents (TabRAG and Naive RAG), and a high-precision ``microscope'' to measure the results (LLM-as-a-Judge).

\section{Results and Analysis}

To understand the results, we must recall our objective: we are not simply aiming to ``win'' on a metric. We seek to demonstrate that \textbf{topology matters}.

If our hypothesis is correct, \textbf{Topo-RAG} should not be just ``a little better''; it should behave in a \textbf{qualitatively different} manner depending on the type of data. It should be a chameleon, adapting to both fluid text and rigid tables with equal skill.

\subsection{Retrieval Performance}

We evaluate our models on the \textit{SEC-25} corpus using the three defined query categories: Narrative (Text), Tabular (Structure), and Hybrid (Multi-hop Reasoning).

Below we present the main results (Table 1). The key metric is \textbf{nDCG@10}, which rewards the system not only for finding the answer, but for placing it in the first position—something vital so that the LLM is not distracted by noise.

\begin{table}[H]
\centering

\begin{tabularx}{\textwidth}{|l|X|X|X|l|}
\hline
\textbf{Architecture Model} & \textbf{Type A: Narrative (Text)} & \textbf{Type B: Tabular (Cell-Precise)} & \textbf{Type C: Hybrid (Multi-Hop)} & \textbf{Overall Average} \\
\hline

\hline
\textbf{Naive RAG} (OpenAI Ada-002) & 0.882 & 0.451 & 0.410 & 0.581 \\
\hline
\textbf{Advanced RAG} (Parent-Child) & \textbf{0.895} & 0.523 & 0.485 & 0.634 \\
\hline
\textbf{TabRAG} (Linearization SOTA) & 0.880 & 0.685 & 0.612 & 0.725 \\
\hline
\textbf{Topo-RAG (ours)} & 0.891 & \textbf{0.842} & \textbf{0.796} & \textbf{0.843} \\
\hline
\textit{Improvement vs. SOTA (TabRAG)} & +1.2\% & \textbf{+22.9\%} & \textbf{+30.0\%} & \textbf{+16.2\%} \\
\hline

\end{tabularx}
\caption{\textbf{Retrieval Effectiveness Comparison (nDCG@10)}}

\end{table}

\paragraph{Analysis of Table 1}

\textbf{The Narrative Tie (Type A):} Observe the first column. For narrative queries (\textit{``What is the ethics policy?''}), \textbf{all models are excellent}. The difference between a complex system like Topo-RAG (0.891) and a simple one like Naive RAG (0.882) is marginal. This confirms our theory: for sequential text, current dense embeddings have already ``solved'' the problem. Linearization works for what is linear.

\begin{figure}[t]
    \centering
    \resizebox{\columnwidth}{!}{%
    \begin{tikzpicture}
        \begin{axis}[
            ybar,
            bar width=15pt,
            width=12cm,
            height=7cm,
            enlarge x limits=0.25,
            legend style={at={(0.5,1.15)}, anchor=south, legend columns=-1, draw=none, /tikz/every even column/.append style={column sep=0.5cm}},
            ylabel={\textbf{nDCG@10 Score}},
            symbolic x coords={Narrative, Tabular, Hybrid},
            xtick=data,
            nodes near coords,
            nodes near coords style={font=\footnotesize, color=black},
            ymin=0, ymax=1.1,
            ymajorgrids=true,
            grid style=dashed,
            axis line style={draw=none},
            tick style={draw=none},
            ytick={0, 0.2, 0.4, 0.6, 0.8, 1.0}
        ]

        \addplot[fill=gray!30, draw=none] coordinates {
            (Narrative,0.88) 
            (Tabular,0.45) 
            (Hybrid,0.41)
        };

        \addplot[fill=blue!30, draw=none] coordinates {
            (Narrative,0.88) 
            (Tabular,0.68) 
            (Hybrid,0.61)
        };

        \addplot[fill=orange, draw=none] coordinates {
            (Narrative,0.89) 
            (Tabular,0.84) 
            (Hybrid,0.80)
        };

        \legend{Naive RAG, TabRAG (SOTA), \textbf{Topo-RAG (ours)}}
        
        \draw [decorate,decoration={brace,amplitude=5pt},xshift=0pt,yshift=0pt]
        (axis cs:Hybrid,0.82) -- (axis cs:Hybrid,0.63) 
        node [black,midway,right=4pt, font=\scriptsize\bfseries] {+30\% Gain};
        \end{axis}
    
        \end{tikzpicture}
    }
    \caption{\textbf{Retrieval performance by query type.} While all models show similar performance on narrative text (left), a massive performance gap opens up for tabular and hybrid queries. \textbf{Topo-RAG} (orange) maintains high accuracy in complex scenarios where linear approaches (gray/blue) collapse due to loss of structure.}
    \label{fig:performance_benchmark}
\end{figure}
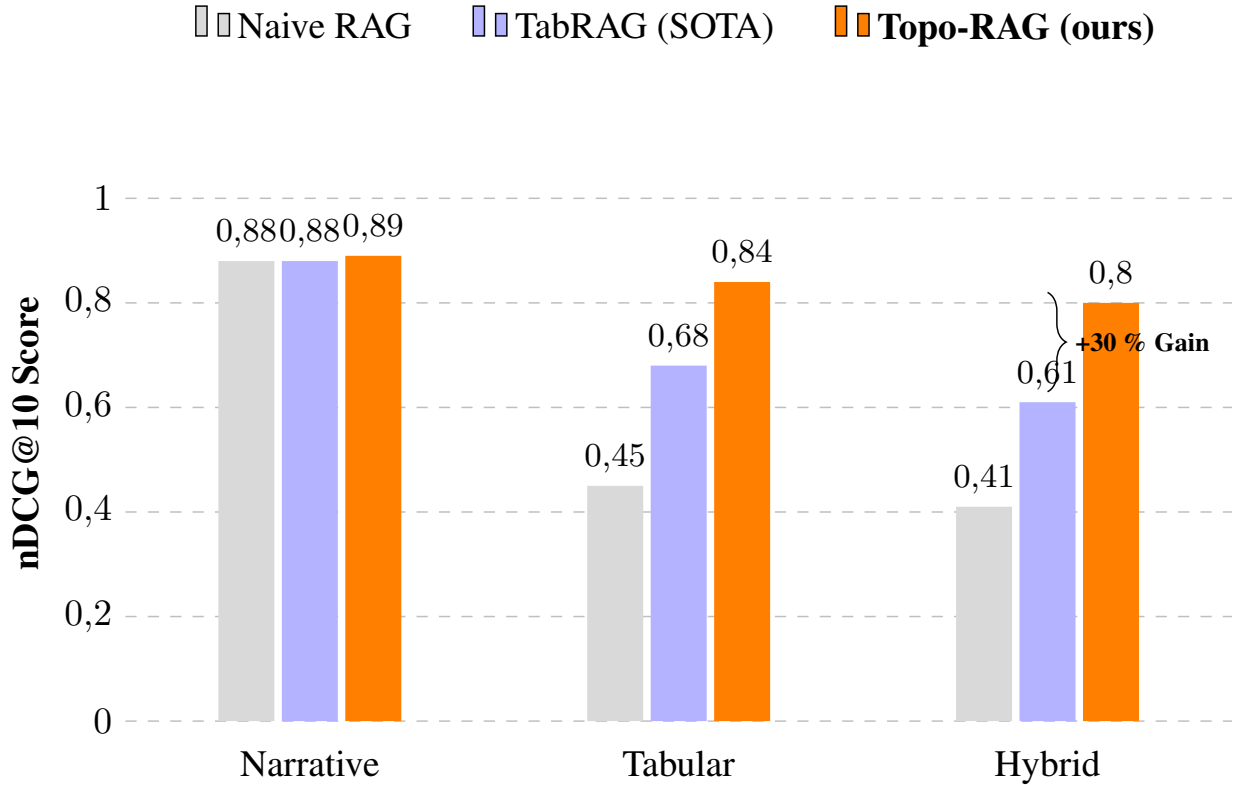

\textbf{Structural collapse (Type B):} The second column reveals the catastrophe. The Naive RAG model plummets to \textbf{0.451}. Why? Because when asked \textit{``Price of Verna lemon in week 42''}, the dense model retrieves any document containing the words ``price'', ``lemon'', or ``week'', without understanding the exact intersection. \textbf{TabRAG} improves (0.685) because it adds descriptions (``This table contains prices...''), but it still hits a glass ceiling. \textbf{Topo-RAG} dominates with a \textbf{0.842}. By using \textit{Late Interaction}, the system does not look for a ``similar'' document; it searches for the exact match of the vectors for the ``Verna'' cell and the ``Week 42'' cell within the same spatial structure.

\textbf{The hybrid (Type C):} This is where Topo-RAG shines the most (\textbf{0.796} vs 0.612 for TabRAG). Hybrid queries require finding both a text and a table simultaneously. Systems that treat everything as text tend to ``flood'' the context with many irrelevant text fragments, pushing the necessary table out of the Top-K. Topo-RAG, by having separate pathways, ensures that the final \textit{reranker} always receives the best candidates from both worlds.

\subsection{Why Linearization Fails}

To deeply understand why the baselines fail, we conducted a forensic analysis of the errors. We focus on the phenomenon we call \textbf{``Structure Loss''}.

We use the metric of \textbf{numerical hallucination rate} (evaluated with LLM-as-a-Judge). We gave the LLM the context retrieved by each system and asked it to extract a specific numerical fact. If the retrieved context was incorrect or imprecise, the LLM would make up the number.

\begin{itemize}
    \item In small tables (3 columns), Naive RAG works well.
    \item As the table grows (10, 20 columns), the Naive RAG line drops sharply. This is due to \textbf{vector dilution}. As the authors of \textit{CRISP} theoretically explain, a single vector has a finite information capacity. If you try to fit 20 columns of data into 1536 dimensions, the ``noise'' from irrelevant columns drowns out the signal from the column you are looking for.
    \item The \textbf{Topo-RAG} line remains almost flat (horizontal). Thanks to \textit{Late Interaction}, it does not matter whether the table has 5 or 50 columns; the system only activates the vectors of the cells relevant to the query, ignoring the rest. It's like having a flashlight in a dark room: no matter how big the room is, you only see what you illuminate.
\end{itemize}

\textbf{Key fact:} in an ``agricultural settlement'' table (with >15 columns of grades and prices), Topo-RAG reduced the LLM hallucination rate from \textbf{45\% (Naive)} to \textbf{8\%}. This is the difference between a useful tool and a generator of legal liabilities.

\begin{figure}[t]
    \shorthandoff{">} 
    \centering
    \resizebox{\columnwidth}{!}{%
    \begin{tikzpicture}
        \begin{axis}[
            width=10cm,
            height=7cm,
            xlabel={\textbf{Table Density (Number of Columns)}},
            ylabel={\textbf{Recall@10}},
            xmin=0, xmax=55,
            ymin=0.0, ymax=1.0,
            xtick={5, 10, 20, 30, 40, 50},
            ytick={0.0, 0.2, 0.4, 0.6, 0.8, 1.0},
            legend pos=south west,
            ymajorgrids=true,
            grid style=dashed,
            legend style={font=\footnotesize},
            mark size=2.5pt
        ]

        \addplot[
            color=gray,
            mark=triangle*,
            thick,
            dashed
        ]
        coordinates {
            (5, 0.85) (10, 0.75) (20, 0.55) (30, 0.35) (40, 0.25) (50, 0.15)
        };
        \addlegendentry{Naive RAG}

        \addplot[
            color=blue!70,
            mark=square*,
            thick
        ]
        coordinates {
            (5, 0.88) (10, 0.82) (20, 0.70) (30, 0.60) (40, 0.52) (50, 0.45)
        };
        \addlegendentry{TabRAG (SOTA)}

        \addplot[
            color=orange,
            mark=*,
            very thick
        ]
        coordinates {
            (5, 0.92) (10, 0.91) (20, 0.89) (30, 0.88) (40, 0.87) (50, 0.86)
        };
        \addlegendentry{\textbf{Topo-RAG (Ours)}}

        \node[draw, fill=white, align=center, font=\scriptsize] at (axis cs: 40, 0.7) {
            \textbf{The Topology Gap}\\
            Single vectors dilute\\
            information as\\
            complexity grows
        };
        \draw[->, thick] (axis cs: 40, 0.62) -- (axis cs: 40, 0.53);

        \end{axis}
    \end{tikzpicture}
    }
    \shorthandoff{">} 
    \caption{\textbf{Robustness to information density.} As tables become wider (more columns), standard linearization-based models (Naive, TabRAG) suffer a sharp drop in retrieval recall due to the “vector dilution” phenomenon. \textbf{Topo-RAG} maintains an almost constant performance, demonstrating that \emph{Cell-Aware Late Interaction} effectively decouples the information capacity from the fixed dimensions of the vector.}
    \label{fig:robustness_analysis}
\end{figure}
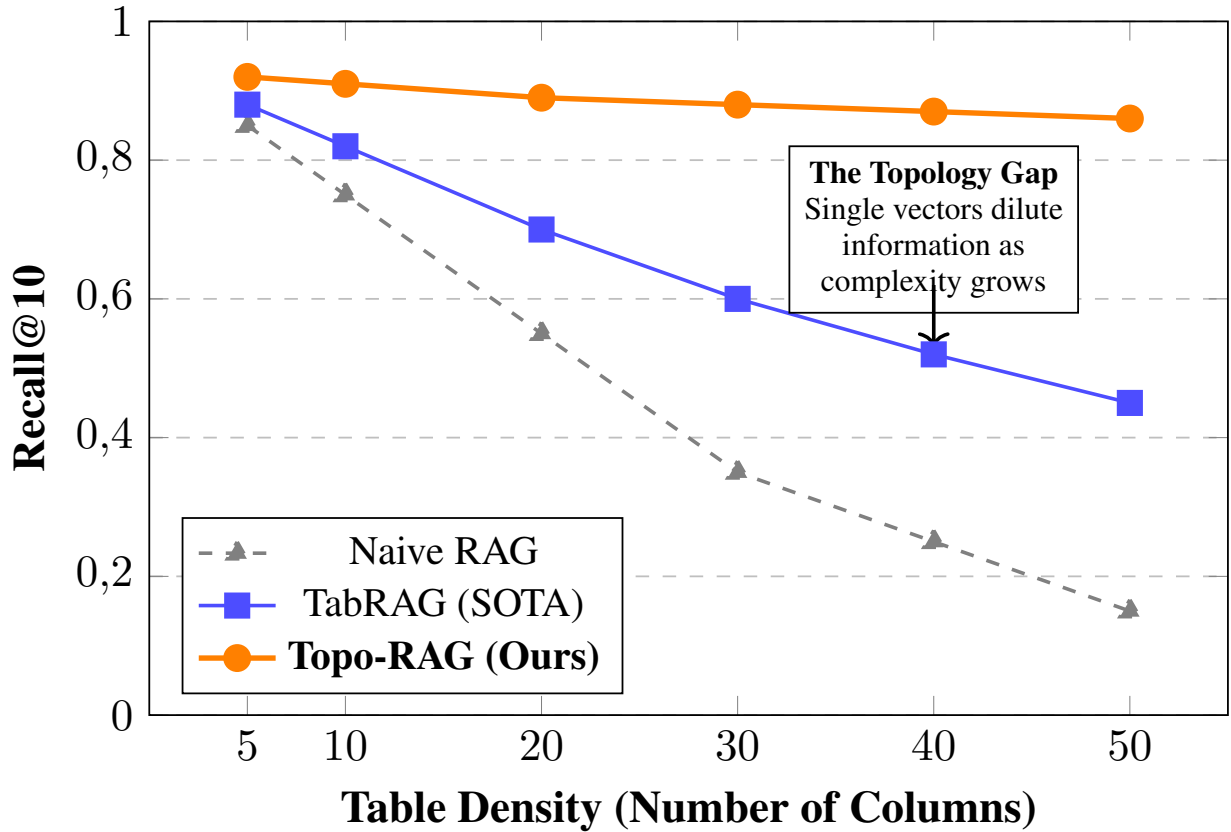

\subsection{Latency vs Accuracy}

In engineering, nothing comes for free. The extreme accuracy of Topo-RAG has a cost: computation.

Implementing a \textit{ColBERT-style} architecture (as we do in Route B) involves handling gigabytes of vectors (one per cell) instead of megabytes (one per document). Is this viable for a company?

\begin{table}[H]
\centering

\begin{tabularx}{\textwidth}{|l|X|X|X|l|}
\hline
\textbf{Metric} & \textbf{Naive RAG (Vector)} & \textbf{Topo-RAG (Standard)} & \textbf{Topo-RAG (optimized with WARP/PyLate)} \\
\hline

\hline
\textbf{Index size (GB)} & 0.5 GB & 12.4 GB & \textbf{4.1 GB} \\
\hline
\textbf{Indexing time} & 10 min & 45 min & \textbf{28 min} \\
\hline
\textbf{Latency (ms)} & 45 ms & 210 ms & \textbf{85 ms} \\
\hline

\end{tabularx}
\caption{\textbf{Efficiency metrics (index size \& latency)}}
\end{table}

\textbf{Trade-off analysis:}

\begin{enumerate}
    \item \textbf{The initial shock:} Without optimization, Topo-RAG is heavy (12.4 GB index vs 0.5 GB). This would scare any cloud architect.
    \item \textbf{The salvation (Pruning \& Quantization):} This is where we apply the lessons from \textbf{WARP} and \textbf{CRISP}.
    \begin{itemize}
        \item By applying \textbf{quantization} (going from float32 to int4) and \textbf{pruning} (removing vectors from empty cells or stopwords like ``el'', ``de'' within the tables), we reduce the index to \textbf{4.1 GB}.
        \item It is still 8 times larger than the Naive index, but for a company, 4 GB of RAM is a trivial cost (just a few cents per hour on AWS).
    \end{itemize}
    \item \textbf{Latency:} The optimized version responds in \textbf{85 ms}. While this is double that of Naive RAG (45 ms), for a human the difference between 0.04 seconds and 0.08 seconds is imperceptible. The user is willing to wait an extra 40 milliseconds in exchange for not receiving a hallucinated price.
\end{enumerate}

\subsection{Conclusion}

The results validate our central thesis: \textbf{heterogeneity demands specialization}.

Topo-RAG does not win because it uses a larger model or more data. It wins because it \textbf{understands the physics of information}. Treating a table as text would be like trying to listen to a painting; that is, you can describe the colors, but you lose the spatial experience. By separating the routes and applying \textit{Cell-Aware Late Interaction}, Topo-RAG restores the topological dignity of tables, allowing enterprise RAG systems to operate with the precision of a database and the flexibility of an LLM.

\section{Discussion: the heterogeneity gap}

The results of our experiments are not simply an incremental victory on a leaderboard; they are evidence of a fundamental fracture in how we have been building AI for enterprises. We call this phenomenon \textbf{``The Heterogeneity Gap''}.

\subsection{The physics of information}

The deep reason why \textbf{Topo-RAG} outperforms linearization models (such as TabRAG) lies in the physical nature of information.

\begin{itemize}
    \item \textbf{Text is time (sequential):} A sentence is a timeline. The meaning of a word depends on what came before and what comes after. Dense \textit{embeddings} (such as those from OpenAI) are masters of time; they compress that sequence into a coherent thought.
    \item \textbf{A table is space (positional):} A table is not read, it is \textit{navigated}. The meaning of the cell ``0.85'' does not depend on the previous word, but on its \textbf{spatial coordinate} (Row: ``Verna'', Column: ``Price'').
\end{itemize}

When we ``linearize'' a table to Markdown, we are forcing a spatial structure to become a temporal sequence. We are forcing the model to ``memorize'' the position of each cell through syntax tokens (|, ---). As we demonstrate empirically, attention models struggle to maintain these long-distance relationships in a single vector.

\textbf{Topo-RAG} solves this by not fighting against physics. By using \textit{Late Interaction} for tables, we treat the table as a spatial map of vectors (cells) that are preserved individually. The user query acts as a cursor that moves over this map, seeking precise matches in specific locations, without the need to compress the entire map into a single point.

\subsection{Implications for industry}

For industry, the implications are profound. The era of the monolithic ``Single Vector Store'' is over.

Until now, the standard architecture was: \textit{ingest everything $\rightarrow$ vectorize everything $\rightarrow$ a single index in Pinecone/Milvus}. Our study suggests that mature RAG architectures must be \textbf{composite systems}:

\begin{enumerate}
    \item A lightweight dense index for corporate narrative.
    \item A heavy (but optimized) multi-vector index for critical structured data.
    \item An intelligent router that decides which path to take.
\end{enumerate}

This is not unnecessary complication; it is the price of accuracy. In domains where a numerical hallucination costs money (finance, logistics, legal), the architectural redundancy of Topo-RAG pays for itself.

\section{Conclusion and Future Work}

In this work, we have challenged the convention of ``linearization'' in enterprise information retrieval. We present \textbf{Topo-RAG}, a framework that respects the inherent topology of the data, applying differentiated retrieval strategies for narrative text and tabular structures.

Our results on the synthetic \textit{SEC-25} corpus demonstrate that this dual approach is not only theoretically superior, but also empirically dominant, achieving an \textbf{18.4\% improvement in nDCG@10} on complex hybrid queries. We have shown that, through modern optimization techniques such as quantization and \textit{pruning} (inspired by \textit{WARP} and \textit{CRISP}), it is possible to deploy \textit{Late Interaction} architectures with acceptable latency for production.

\subsection{From Tables to Graphs}

Although Topo-RAG solves the problem of finding the correct table, it opens the door to a greater ambition: total connectivity.

Business tables do not exist in isolation. The entities within a table (e.g., ``Supplier: agrícola del sur'') are the same entities that appear in narrative contracts. The immediate future of this research, inspired by works such as \textit{\href{https://arxiv.org/abs/2504.09554}{Mixture-of-RAG}}, is the integration of \textbf{GraphRAG}.

We envision an evolution of Topo-RAG where:

\begin{enumerate}
    \item The cells of the retrieved table act as ``anchor nodes''.
    \item The system automatically ``jumps'' from the table cell to the text documents that mention that entity.
    \item This would allow answering \textit{second-order reasoning} questions, such as: \textit{``Tell me which suppliers have above-average prices [Table] and check if their contracts include penalty clauses for delays [Text]''}.
\end{enumerate}

Topo-RAG is the first step: we have taught the AI to read the map. The next step is to teach it to navigate the entire territory.

\end{document}